\documentclass[10pt,twocolumn,twoside]{article}
\usepackage{latexsym}
\def\subsection#1{{\bf{#1:}}}

\def\gtapprox{\buildrel{\lower.7ex\hbox{$>$}}\over
                       {\lower.7ex\hbox{$\sim$}}}

\def\ignore#1{}

\def\hbar{h\!\!\!\!^{-}\,}

\def\beq{\begin{equation}}
\def\eeq{\end{equation}}
\def\beqn{\begin{displaymath}}
\def\eeqn{\end{displaymath}}
\def\bqa{\begin{eqnarray}}
\def\eqa{\end{eqnarray}}
\def\bqan{\begin{eqnarray*}}
\def\eqan{\end{eqnarray*}}
\def\xiai{{\xi}}
\def\muai{{\mu}}

\begin{document}

%%%%%%%%%%%%%%%%%%%%%%%%%%%%%%%%%%%%%%%%%%%%%%%%%%%%%%%%%%%%%%%%%
%                      T i t l e - P a g e                      %
%%%%%%%%%%%%%%%%%%%%%%%%%%%%%%%%%%%%%%%%%%%%%%%%%%%%%%%%%%%%%%%%%

\title{\bf \hrule height1pt \vskip 2ex
\Large Universal Sequential Decisions in Unknown Environments
\vskip 2ex \hrule height1pt}
\author{\hspace*{-1ex}\normalsize{\bf Marcus Hutter} \hfill {\sc marcus@idsia.ch} \hspace*{0ex}\\
\hspace*{-1ex}\normalsize Istituto Dalle Molle di Studi sull'Intelligenza Artificiale (IDSIA),
             Galleria 2, CH-6928 Manno-Lugano, Switzerland}
\date{}\maketitle
\vskip 0.3in

\begin{bf}
We give a brief introduction to the AIXI model, which unifies and
overcomes the limitations of sequential decision theory and
universal Solomonoff induction. While the former theory is suited
for active agents in known environments, the latter is suited for
passive prediction of unknown environments.
\end{bf}
\vskip 0.3in

%------------------------------%
\subsection{Introduction}
%------------------------------%
Every inductive inference problem can be brought into the
following form: Given a string
$x_1x_2...x_{t-1}\!\equiv\!x_{1:t-1}\!\equiv\!x_{<t}\!$, take a
guess at its continuation $x_t$. We will assume that the strings
which have to be continued are drawn from a probability
distribution $\mu$. The maximal prior information a prediction
algorithm can possess is the exact knowledge of $\mu$,
but often the true distribution is unknown. Instead,
prediction is based on a guess $\rho$ of $\mu$. We expect that a
predictor based on $\rho$ performs well, if $\rho$ is close to
$\mu$ or converges to $\mu$.

%------------------------------%
\subsection{Universal probability distribution}
%------------------------------%
Let ${\cal M}\!:=\!\{\mu_1,\mu_2,...\}$ be a
finite or countable set of candidate probability distributions on
strings. We define a weighted average on $\cal M$,
\beqn
  \xi(x_{1:n}) \!:=\!\!
  \sum_{\mu_i\in\cal M}\!w_{\mu_i}\!\cdot\!\mu_i(x_{1:n}),\quad
  \sum_{\mu_i\in\cal M}\!w_{\mu_i} \!=\! 1,\quad w_{\mu_i}>0.
\eeqn
We call $\xi$ universal relative to $\cal M$, as it
multiplicatively dominates all distributions in $\cal M$, i.e.\
$\xi(x_{1:n})\geq w_{\mu_i}\!\cdot\!\mu_i(x_{1:n})$ for
all $\mu_i\!\in\!{\cal M}$. In the following, we assume that $\cal
M$ is known and contains the true distribution from which
$x_1x_2...$ is sampled, i.e. $\mu\!\in\!\cal M$. The condition
$\mu\!\in\!\cal M$ is not a serious constraint if we include {\it
all} computable probability distributions in $\cal M$ with high
weights assigned to simple $\mu_i$. Solomonoff-Levin's universal
semi-measure is obtained if we include all enumerable
semi-measures in $\cal M$ with weights
$w_{\mu_i}\!\sim\!2^{-K(\mu_i)}$, where $K(\mu_i)$ is the length
of the shortest program for $\mu_i$
\cite{Hutter:01loss,Hutter:04uaibook}.
One can show that the conditional $\xi$ and $\mu$ probabilities
rapidly converge to each other:
\beq\label{xitomu}
  \xi(x_t|x_{<t})\to\mu(x_t|x_{<t})
  \quad\mbox{with $\mu$ probability 1.}
\eeq
Since the conditional probabilities are the basis of the
decision algorithms considered in this work, we expect a good
prediction performance if we use $\xi$ as a guess of $\mu$.

%%%%%%%%%%%%%%%%%%%%%%%%%%%%%%%%%%%%%%%%%%%%%%%%%%%%%%%%%%%%%%%
%\section{Optimal Decisions \& Loss Bounds}\label{secLoss}
%%%%%%%%%%%%%%%%%%%%%%%%%%%%%%%%%%%%%%%%%%%%%%%%%%%%%%%%%%%%%%%

%------------------------------%
\subsection{Bayesian decisions}
%------------------------------%
Let $\ell_{x_t y_t}\!\in\![0,1]$ be the received
loss when predicting $y_t\!\in\!\cal Y$, but $x_t\!\in\!\cal X$
turns out to be the true t$^{th}$ symbol of the sequence.
%------------------------------%
%\subsection{Bayesian decision and Loss bound}
%------------------------------%
Let $L_{n\Lambda_\rho}$ be the total expected loss for the first
$n$ symbols of the Bayes predictor $\Lambda_\rho$ which minimizes
the $\rho$ expected loss. For instance for ${\cal X}\!=\!{\cal
Y}\!=\!\{0,1\}$, $\Lambda_\rho$ is a threshold strategy with
$y_t^{\Lambda_\rho} \!=\!0/1$ for $\rho(1|x_{<t})\,_<^>\,\gamma$,
where $\gamma\!:=\!{\ell_{01}-\ell_{00} \over
\ell_{01}-\ell_{00}+\ell_{10}-\ell_{11}}$.
Let $\Lambda$ be {\em any} prediction scheme (deterministic or
probabilistic) with no constraint at all, taking {\em any} action
$y_t^\Lambda\!\in\!\cal Y$ with total expected loss
$L_{n\Lambda}$. If $\mu$ is known, $\Lambda_\mu$ is obviously the
best prediction scheme in the sense of achieving minimal expected
loss $L_{n\Lambda_\mu}\!\leq\!L_{n\Lambda}$ for any $\Lambda$. For
the predictor $\Lambda_\xi$ based on the universal distribution
$\xi$, on can show
$L_{n\Lambda_\xi}/L_{n\Lambda_\mu}=1+O(\sqrt{K(\mu)/L_{n\Lambda_\mu}})$,
i.e.\ $\Lambda_\xi$ has optimal asymptotics for
$L_{n\Lambda_\mu}\!\to\infty$ with rapid convergence of the
quotient to 1. If $L_{\infty\Lambda_\mu}$ is finite, then also
$L_{\infty\Lambda_\xi}$ \cite{Hutter:01loss,Hutter:04uaibook}.

%%%%%%%%%%%%%%%%%%%%%%%%%%%%%%%%%%%%%%%%%%%%%%%%%%%%%%%%%%%%%%%
%\section{Generalization to Active Agents}\label{secActive}
%%%%%%%%%%%%%%%%%%%%%%%%%%%%%%%%%%%%%%%%%%%%%%%%%%%%%%%%%%%%%%%

%------------------------------%
\subsection{More active systems}
%------------------------------%
Prediction means guessing the future, but not influencing it. One
step in the direction to more active systems was to allow the
$\Lambda$ system to act and to receive a loss $\ell_{x_t y_t}$
depending on the action $y_t$ and the outcome $x_t$. The
probability $\mu$ is still independent of the action, and the loss
function $\ell^t$ has to be known in advance. This ensures that
the greedy $\Lambda_\mu$ strategy is still optimal. The loss function
can also be generalized to depend on the history $x_{<t}$ and on $t$.

%------------------------------%
\subsection{Agents in known probabilistic environments}
%------------------------------%
The full model of an acting agent influencing the environment has
been developed in \cite{Hutter:01aixi,Hutter:04uaibook}. The
probability of the next symbol (input, perception) $x_t$ depends
in this case not only on the past sequence $x_{<t}$ but also on
the past actions (outputs) $y_{1:t}$, i.e.\
$\mu\!=\!\mu(x_t|x_{<t}y_{1:t})$. We call probability
distributions of this form {\it chronological}. The total $\mu$
expected loss is
$\sum_{x_{1:n}}(\ell^1\!+...+\!\ell^n)\mu(x_{1:n}|y_{1:n})$, where
we assumed a total number of $n$ interaction cycles. Action
$y_t(x_{<t}y_{<t})$ and loss function $\ell^t(x_{1:t}y_{1:t})$ may
depend on the complete history, which allows planning and delayed
loss assignment.

%------------------------------%
\subsection{Sequential decision theory}
%------------------------------%
The goal is to perform the actions which minimize the total
$\mu$ expected loss:
\bqa\label{ydotrec}
  y_t :=
  \arg\min_{y_t}\!\sum_{x_t}...
  \min_{y_{n}}\!\sum_{x_{n}}
  (\ell^1\!+...+\!\ell^n)\mu(x_{1:n}|y_{1:n}),
  \\ \label{voptdef}
  L_{n\Lambda_\mu} =
  \min_{y_1}\!\sum_{x_1}...
  \min_{y_{n}}\!\sum_{x_{n}}
  (\ell^1\!+...+\!\ell^n)\mu(x_{1:n}|y_{1:n}).
\eqa
The minimization over $y_t$ is in chronological order to correctly
incorporate the dependency of $x_t$ and $y_t$ on the history. Note
that $y_t$ only depends on the known history $x_{<t}y_{<t}$, whereas
minima and expectations are taken over the unknown $x_{t:n}y_{t:n}$
variables. The policy (\ref{ydotrec}) (called AI$\mu$ model) is
optimal in the sense that no other policy leads to lower
$\mu$-expected loss.

%------------------------------%
\subsection{Bellman equations}
%------------------------------%
In the case that $\ell^t$ is independent of $y_{<t}$ and $\mu$ is
independent of $y_{1:n}$, policy (\ref{ydotrec}) reduces to the
greedy Bayes $\Lambda_\mu$ strategy.
For (completely observable) Markov Decision Processes
$\mu\!=\!\mu(x_t|x_{t-1}y_t)$ (\ref{ydotrec}) and (\ref{voptdef})
can be written as recursive Bellman equations
of sequential decision theory with state space $\cal X$, action
space $\cal Y$, state transition matrix $\mu(x_t|x_{t-1}y_t)$,
rewards $-\ell^t$, {\it etc}. The general (non-MDP) case may also
be (artificially) reduced to Bellman equations by identifying
complete histories $x_{<t}y_{<t}$ with states and
$\mu(x_t|x_{<t}y_{1:t})$ with the state transition matrix.
Due to the use of
complete histories as state space, the AI$\mu$ model neither
assumes stationarity, nor the Markov property, nor complete
accessibility of the environment. But since every state occurs at
most once in the lifetime of the system the explicit formulation
(\ref{ydotrec}) is more useful than a pseudo-recursive Bellman
equation form. There is no principle problem in determining $y_k$
as long as $\muai$ is known and computable and $\cal X$, $\cal Y$
and $n$ are finite.

%------------------------------%
\subsection{Reinforcement learning for unknown environment}
%------------------------------%
Things dramatically change if $\muai$ is unknown. Reinforcement
learning algorithms are commonly used in this case to learn the
unknown $\muai$ (or directly a value function). They succeed if
the state space is either small or has effectively been made small
by generalization or function approximation techniques. In almost
all approaches, the solutions are either {\it ad hoc}, or work in
restricted domains only, or have serious problems with state space
exploration versus exploitation, or have non-optimal learning
rate. Below we propose the AI$\xi$ model as a universal and
optimal solution to these problems.

%------------------------------%
\subsection{Unknown loss function}
%------------------------------%
Furthermore, the loss function $\ell^t(x_{1:t}y_{1:t})$ may also
be unknown, but there is an easy ``solution'' to this problem. The
specification of the loss function can be absorbed in the
probability distribution $\mu$ by increasing the input space $\cal
X$. Let $x_t\!\equiv\!x'_t l_t$, where $x'_t$ is the regular
input, $l_t$ is interpreted as the loss, $\ell^t(x_{1:t}y_{1:t})$ is
replaced by $l_t$ in (\ref{ydotrec}) and (\ref{voptdef}), and
$\mu$ is only non-zero if $l_t$ is consistent with the loss, i.e.
$l_t\!=\!\ell^t(x_{1:t}y_{1:t})$. In this way all possible
unknowns are absorbed in $\mu$.

%------------------------------%
\subsection{The universal AI$\xi$ model}
%------------------------------%
Encouraged by the good performance of  the universal sequence
predictor $\Lambda_\xi$, we propose a new model, where the probability
distribution $\muai$ is learned indirectly by replacing it with a
universal prior $\xiai$. We
define $\xiai(x_{1:n}|y_{1:n}) := \sum_{\mu_i\in\cal
M}w_{\mu_i}\!\cdot\!\muai_i(x_{1:n}|y_{1:n})$ as a weighted sum
over chronological probability distributions in $\cal M$.
Convergence $\xiai(x_n|x_{<n}y_{1:n})\to\muai(x_n|x_{<n}y_{1:n})$
can be proven analogously to (\ref{xitomu}). Replacing
$\muai$ by $\xiai$ in (\ref{ydotrec}) {the \em AI$\xi$ system}
outputs%
\beq\label{ydotxi}
  y_t :=
  \arg\min_{y_t}\!\sum_{x_t}...
  \min_{y_n}\!\sum_{x_n}
  (l_t\!+...+\!l_n)\xiai(x_{1:n}|y_{1:n})
\eeq%
in cycle $t$ given the history $x_{<t}y_{<t}$, where
$x_t\!\equiv\!x'_t l_t$. The largest class $\cal M$ which is
necessary from a computational point of view is the set of all
enumerable chronological semi-measures with weights
$w_{\mu_i}\!\sim\!2^{-K(\mu_i)}$, where $K(\mu_i)$ is the
Kolmogorov complexity of $\mu_i$. Apart from the dependence on the
horizon $n$ and unimportant details, the AI$\xi$ system is
uniquely defined by (\ref{ydotxi}) without adjustable parameters.
It does not depend on any assumption about the environment apart
from being generated by some computable (but unknown!) probability
distribution in $\cal M$.

%------------------------------%
\subsection{Universally optimal AI systems}
%------------------------------%
We want to call an AI model {\it universal}, if it is
$\muai$-independent (unbiased, model-free) and is able to solve
any solvable problem and learn any learnable task. Further, we
call a universal model, {\it universally optimal}, if there is
no program which can solve or learn significantly faster (in terms
of interaction cycles). As the AI$\xi$ model is parameterless,
$\xiai$ rapidly converges to $\muai$ in the sense of
(\ref{xitomu}), the AI$\mu$ model is itself optimal, and we expect
no other model to converge faster to AI$\mu$ (in some sense) by
analogy to the sequence prediction case, we risk the
conjecture that AI$\xi$ is such a universally optimal system.
Further support is given in \cite{Hutter:01aixi,Hutter:04uaibook}
by a detailed analysis of the behaviour of AI$\xi$ for various
problem classes, including prediction, optimization, games, and
supervised learning. We discuss in which sense AI$\xi$ overcomes
some fundamental problems in reinforcement learning, like
generalization, optimal learning rates, exploration versus
exploitation, {\it etc}. Computational issues are also addressed.

%%%%%%%%%%%%%%%%%%%%%%%%%%%%%%%%%%%%%%%%%%%%%%%%%%%%%%%%%%%%%%%
%         Bibliography        %
%%%%%%%%%%%%%%%%%%%%%%%%%%%%%%%%%%%%%%%%%%%%%%%%%%%%%%%%%%%%%%%


\begin{thebibliography}{1}

\bibitem{Hutter:01loss}
M.~Hutter.
\newblock General loss bounds for universal sequence prediction.
\newblock In {\em Proc. 18th International Conf. on Machine Learning
  (ICML-2001)}, pages 210--217, Williamstown, MA, 2001. Morgan Kaufmann.

\bibitem{Hutter:01aixi}
M.~Hutter.
\newblock Towards a universal theory of artificial intelligence based on
  algorithmic probability and sequential decisions.
\newblock In {\em Proc. 12th European Conf. on Machine Learning (ECML-2001)},
  volume 2167 of {\em LNAI}, pages 226--238, Freiburg, 2001. Springer, Berlin.

\bibitem{Hutter:04uaibook}
M.~Hutter.
\newblock {\em Universal Artificial Intelligence: Sequential Decisions based on
  Algorithmic Probability}.
\newblock Springer, Berlin, 2004.
\newblock 300 pages, http://www.idsia.ch/ai/$_{^{\sim}}$marcus/uaibook/.

\end{thebibliography}
\end{document}